\pdfoutput=1

\documentclass[11pt]{article}

\usepackage[final]{acl}

\usepackage{times}
\usepackage{latexsym}

\usepackage[T1]{fontenc}

\usepackage[utf8x]{inputenc}
\usepackage{ucs}

\usepackage{microtype}

\usepackage{inconsolata}
\usepackage{soul}
\usepackage{graphicx}
\usepackage{listings}

\definecolor{codegreen}{rgb}{0,0.6,0}
\definecolor{codegray}{rgb}{0.5,0.5,0.5}
\definecolor{codepurple}{rgb}{0.58,0,0.82}
\definecolor{backcolour}{rgb}{0.95,0.95,0.92}
\title{M-DAIGT: A Shared Task on Multi-Domain Detection of AI-Generated Text}

\author{
  Salima Lamsiyah\textsuperscript{1},
  Saad Ezzini\textsuperscript{2},
  Abdelkader {El Mahdaouy}\textsuperscript{3},
  Hamza Alami\textsuperscript{4},\\
  \textbf{Abdessamad Benlahbib\textsuperscript{4}},
  \textbf{Samir El Amrany\textsuperscript{1}},
  \textbf{Salmane Chafik\textsuperscript{3}},
  \textbf{Hicham Hammouchi\textsuperscript{1}}
  \\
  \\
  \textsuperscript{1}University of Luxembourg, Luxembourg\\
  \textsuperscript{2}King Fahd University of Petroleum and Minerals, Saudi Arabia\\
  \textsuperscript{3}Mohammed VI Polytechnic University, Morocco\\
  \textsuperscript{4}Sidi Mohamed Ben Abdellah University, Morocco\\
  \\
}

\begin{document}
\maketitle
\begin{abstract}
The generation of highly fluent text by Large Language Models (LLMs) poses a significant challenge to information integrity and academic research. In this paper, we introduce the Multi-Domain Detection of AI-Generated Text (M-DAIGT) shared task, which focuses on detecting AI-generated text across multiple domains, particularly in news articles and academic writing. M-DAIGT comprises two binary classification subtasks: News Article Detection (NAD) (Subtask 1) and Academic Writing Detection (AWD) (Subtask 2). To support this task, we developed and released a new large-scale benchmark dataset of 30,000 samples, balanced between human-written and AI-generated texts. The AI-generated content was produced using a variety of modern LLMs (e.g., GPT-4, Claude) and diverse prompting strategies. A total of 46 unique teams registered for the shared task, of which four teams submitted final results. All four teams participated in both Subtask 1 and Subtask 2. We describe the methods employed by these participating teams and briefly discuss future directions for M-DAIGT.
\end{abstract}

\section{Introduction}

The recent advancements in large language models have created a paradigm shift in content generation \cite{naveed2023comprehensive,chang2024survey}. These models offer numerous opportunities to improve a wide range of applications, including academic research and journalism \cite{chung-etal-2023-increasing}. However, their powerful capabilities also raise critical concerns regarding the integrity of the information ecosystem \cite{wu2025survey}. In journalism, the potential for large-scale automated generation of misinformation and fake news represents a serious societal threat, with AI-generated articles already appearing on both mainstream and disinformation websites \cite{wu2025survey, ali-etal-2025-detection}. In academia, LLMs challenge the fundamental principles of academic honesty \cite{bittle2025generative}, and the accessibility of these tools has made it easier for students to generate ghostwritten assignments, contributing to a noticeable rise in academic misconduct \cite{bittle2025generative, go-etal-2025-xdac}. Research indicates that a significant number of students acknowledge using such tools for their coursework, making it increasingly difficult to distinguish between appropriate academic support and plagiarism \cite{kovari2025ethical}.

Distinguishing AI-generated text from human writing is a non-trivial scientific challenge. Modern LLMs produce text that is grammatically correct, stylistically coherent, and often factually plausible, making it difficult to differentiate from human output \cite{brown2020language,urlana-etal-2024-trustai, mitchell2023detectgpt}. Empirical studies have shown that humans, including experienced educators with high confidence in their judgment, perform only marginally better than random chance when attempting to distinguish AI-generated text from human-written content \cite{urlana-etal-2024-trustai}. Moreover, recent detection approaches, such as entropy-based statistical methods \cite{shen2023textdefense}, syntactic pattern analysis \cite{tassopoulou2021enhancing}, and neural classifiers \cite{ippolito-etal-2020-automatic, li-etal-2025-iron}, show promise yet remain vulnerable to paraphrasing and prompt variation \cite{rivera-soto-etal-2025-mitigating, kirchenbauer2023watermark}. The field is effectively locked in a technological "arms race": as detection tools improve, so do generative models and the methods used to evade them, including paraphrase attacks and text "humanizers" \cite{wu2025survey, sadasivan2023can}.

Therefore, this rapidly evolving landscape underscores the need for ongoing research and rigorous evaluation methods for AI content detection. The motivation for advancing detection methodologies extends beyond a reactive approach aimed solely at identifying academic dishonesty. Rather, it serves as a proactive strategy to preserve the integrity of the digital information ecosystem. One key concern is the phenomenon of recursive degradation, where future language models may be trained on vast amounts of unlabeled AI-generated text collected from the internet. This process risks diminishing the quality, originality, and diversity of training data, potentially leading to a degradation of model performance over time \cite{wang-etal-2024-m4}. Given that news articles and academic publications constitute essential sources of high-quality training data, maintaining their authenticity is crucial for ensuring the long-term robustness, reliability, and generalization capabilities of future AI systems.

To address some of these challenges and to further encourage work on AI-generated text detection, we organized the \textbf{Multi-Domain Detection of AI-Generated Text (M-DAIGT)} shared task. M-DAIGT focuses on two domains where the authenticity of text is particularly vital: news articles and academic writing. Specifically, the task is structured into two binary classification subtasks:
\begin{itemize}
    \item \textbf{Subtask 1. News Article Detection (NAD):} Classifying news content as human-written or AI-generated.
    \item \textbf{Subtask 2. Academic Writing Detection (AWD):} Classifying academic texts as human-written or AI-generated.
\end{itemize}

The key contributions of this work are as follows: (1) the creation and public release of a large and diverse dataset of 30,000 samples specifically designed for AI-generated text detection in the domains of news and academia, featuring outputs from models like GPT-4 and Claude using varied prompts \cite{wang-etal-2024-m4}; and (2) a comprehensive analysis of participating systems that range from statistical methods to transformer-based detectors \cite{li-etal-2025-iron, kuznetsov-etal-2025-feature}, offering insights into the current state-of-the-art and highlighting key challenges for future research.

The remainder of this paper is organized as follows: Section~\ref{relatedwork} reviews related work in AI-generated text detection. Section~\ref{datasetcollection} presents the dataset creation process and evaluation metrics. Section~\ref{models} presents the baseline and participant models, along with the evaluation methodology and results. Finally, Sections~\ref{conclusion} and~\ref{limitations} conclude the paper and discuss the limitations of the shared task.

\section{Related Work}\label{relatedwork}
\paragraph{AI-Generated Text Detection Methods.}  The detection of AI-generated text is a rapidly evolving research domain, with increasing attention due to the widespread development of large language models \cite{wu2025survey}.  Several methods have been proposed for AI-generated text detection, which can be broadly classified into statistics-based methods, neural-based methods, watermarking, and the use of LLMs as detectors.

\textbf{\textit{Statistics-based approaches}} aim to exploit intrinsic differences in linguistic features between human and machine-generated texts. Early efforts, such as those \citet{shen2023textdefense} and \citet{tassopoulou2021enhancing}, leveraged entropy measures and n-gram frequency analysis to differentiate between text origins. \citet{krishna-etal-2022-rankgen} utilized sentence repetition patterns, noting that LLMs often assign high probability to repetitive content. DetectGPT \cite{mitchell2023detectgpt} proposed a perturbation-based method to identify whether text lies in negatively curved regions of the log-likelihood landscape. 

\textbf{\textit{Neural-based methods}} dominate recent advances in AI-generated text detection due to their high accuracy and adaptability. Early methods adopt fine-tuned models like BERT \cite{devlin-etal-2019-bert} and RoBERTa \cite{solaiman2019release}. Furthermore,  \citet{ippolito-etal-2020-automatic} demonstrated that training on outputs generated using diverse decoding strategies (e.g., top-k sampling, nucleus sampling, temperature control) significantly improves detection robustness.  Recently, \citet{li-etal-2025-iron} proposed IRON, a robust adversarial training framework that improves resilience against attacks designed to evade detection systems. \citet{jiao-etal-2025-rangedetector} introduced M-RangeDetector, which enhances model generalization via multi-range attention masks. Similarly, \citet{kuznetsov-etal-2025-feature} provided feature-level interpretability through sparse autoencoders, offering insights into which patterns distinguish AI and human text. \citet{tong-etal-2025-generate} combined reinforced sampling with LLM augmentation for improved fake news detection, while \citet{ali-etal-2025-detection} extended neural classifiers to low-resource languages, specifically addressing Urdu fake news detection. These efforts reflect a growing focus on robustness, explainability, and multilingual applicability in neural detection research. 

\textbf{\textit{Watermarking-based approaches}} offer proactive detection capabilities by embedding or identifying implicit signals in generated text. Early methods include synonym replacement, lexical substitution \cite{li2023protecting, sadasivan2023can}, and soft watermarking using curated token lists \cite{kirchenbauer2023watermark}. Hidden-space watermarking approaches \cite{zhao2023protecting} manipulate token-level probability vectors to introduce tamper-resistant signatures. Some methods, like \citet{bhattacharjee2024fighting}, aim to exploit surface-level word randomness as a trigger for detection.
Recently, \citet{rivera-soto-etal-2025-mitigating} proposed Paraphrase Inversion, a novel technique to counter paraphrase attacks that aim to remove watermark signals by recovering semantic intent. This approach highlights the challenges posed by adversaries seeking to bypass detection through surface-level text alterations. While many watermarking techniques rely on controlled generation, this method contributes a defensive post-processing solution that does not depend on direct access to generation mechanisms.

Lastly, LLMs themselves are increasingly used as detectors. Tools such as GPTZero\footnote{https://gptzero.me/}, ZeroGPT\footnote{https://www.zerogpt.com/}, and OpenAI’s\footnote{https://platform.openai.com/ai-text-classifier} AI text classifier exemplify this trend. \citet{sadasivan2023can} proposed a zero-shot framework using clustering to differentiate between watermarked and unwatermarked text. \citet{wang-etal-2024-m4} proposed M4, a comprehensive black-box framework for machine-generated text detection that operates across multiple generators, domains, and languages. Their approach focuses on generalization under realistic, diverse conditions by evaluating detectors on unseen generators and multilingual datasets, setting a new benchmark for robust and scalable AI text detection. More recently, \citet{su-etal-2025-haco} introduced HACo-Det, which focuses on fine-grained detection of human-AI coauthored text, a challenging scenario due to subtle stylistic blending. \citet{go-etal-2025-xdac} proposed XDAC, a detection and attribution framework using explainable AI for Korean-language content. \citet{li-wan-2025-writes} examined how authorial intent and role influence AI-text detectability, emphasizing the social and cognitive dimensions of authorship. These works represent a trend toward leveraging LLMs as meta-models that interpret, explain, and critique textual content.

\paragraph{Benchmark Datasets and Shared Tasks.} Standardized evaluation frameworks through both benchmark datasets and shared tasks play a crucial role in advancing AI-generated text detection by providing standardized, diverse, and challenging evaluation settings. A variety of benchmarks have been proposed to test generalization across languages, domains, modalities, and attack scenarios. The MultiSocial dataset \cite{macko-etal-2025-multisocial} supports multilingual detection on social media content, while XDAC \cite{go-etal-2025-xdac} introduces explainable detection and attribution for LLM-generated news comments in Korean. Double Entendre \cite{frohmann-etal-2025-double} expands detection tasks beyond pure text through a multimodal benchmark focused on audio-based AI-generated lyrics. To assess robustness under adversarial conditions, IRON \cite{li-etal-2025-iron} incorporates adversarially perturbed examples, and Stress-Testing \cite{pedrotti-etal-2025-stress} manipulates LLM writing styles to mislead detectors. In parallel, feature-level datasets \cite{kuznetsov-etal-2025-feature} offer interpretable benchmarks using sparse autoencoders, while M4GT-Bench \cite{wang-etal-2024-m4} evaluates black-box detectors across multiple generators, domains, and languages, which is critical for real-world deployment. Additional public resources, such as the \textit{AI-and-Human-Generated-Text} dataset available on Hugging Face\footnote{\url{https://huggingface.co/datasets/Ateeqq/AI-and-Human-Generated-Text}} and the \textit{GPT-generated Text Detection: Benchmark Dataset and Tensor-based Detection Method} \cite{qazi2024gpt}, further enrich the landscape of available datasets.

Complementing benchmark datasets, shared tasks have emerged as key drivers of progress in AI-generated text detection by offering standardized, competitive, and collaborative evaluation platforms. The SemEval-2024 Task 8 \cite{wang-etal-2024-semeval-2024} was specifically designed to evaluate detection systems under multimodal, multidomain, and multilingual settings in a black-box scenario, challenging participants to detect text generated by unseen language models across diverse languages and content types. The task highlighted major real-world concerns such as domain shift, lack of training-time transparency, and linguistic variability.
Among the participating systems, TrustAI \cite{urlana-etal-2024-trustai} provided a comprehensive analysis of multi-domain machine-generated text detection techniques, implementing various strategies across statistical, neural, and ensemble approaches. Their findings underscored the importance of domain-specific fine-tuning and robust feature extraction in black-box detection contexts. In parallel, the 1st Workshop on GenAI Content Detection (GenAIDetect) \cite{genaidetect-ws-2025-1}, held at COLING 2025, provided a dedicated forum for advancing research on generative content detection. It addressed key challenges such as multilingual robustness, adversarial evasion, and watermarking techniques, fostering discussion around emerging benchmarks and methodological innovation. Building on prior efforts, M-DAIGT focused on AI-generated text detection in two critical domains: news journalism and academic writing. It features two binary classification subtasks, News Article Detection (NAD) and Academic Writing Detection (AWD), supported by a newly released dataset. 

\section{Datasets and Evaluation Metrics} \label{datasetcollection}

\subsection{Datasets Collection}
To support the M-DAIGT shared task, we curated a dataset tailored to evaluate systems on detecting AI-generated news and academic texts. 

\subsubsection{News dataset:} We gathered 7,000 manually written news articles from the CNN Daily News website, covering more than 40 categories. To create the AI-generated counterparts, we used the titles extracted from these human-written articles as input prompts. Multiple language models were employed to generate news content, including LLaMA3.2-3B-Instruct, Qwen2.5-3B-Instruct, Mistral-7B-Instruct-v2.0, and various models from the GPT family (GPT-4o, GPT-3.5, GPT-4o-mini). Each model was prompted using the standardized prompt shown in Listing~\ref{lst:news_prompt}, with the role definition randomly selected at runtime to encourage stylistic diversity in the generated outputs.

\begin{figure}
\begin{lstlisting}[caption={Prompt's Key Components for Generating News Articles}, label={lst:news_prompt}]
-- Each time this prompt is used, a role is randomly selected to influence the assistant writing style.

-- Randomly select one of the following journalist roles:

Role Definition: 
    - "You are an expert journalist."
    - "You are a professional news writer with a focus on clear, unbiased reporting."
    - "You are a friendly and engaging journalist, writing in an easy-to-understand style."
    - "You are an opinion writer, focusing on offering personal insights on current news."

-- Instructions:

Generate a news article of approximately '{article_length}'-words on the following topic: '{Title}'

Write only the article content. Do not include a title or any additional commentary.
\end{lstlisting}
\end{figure}

\begin{figure}
\begin{lstlisting}[caption={Prompts for Generating Scientific Abstracts}, label={lst:abstracts_prompt}]
-- Prompt 1:

You are a researcher working on a research paper. Your English proficiency level is '{english_proficiency}'.
Your task is to write a well-structured abstract of approximately 250 words for your research paper in response to the given topic: '{title}'.
Ensure your abstract is clear and concise, following the standard format: 'background', 'objective', 'methodology', 'key findings', and 'conclusion'.
The response should contain only the abstract text, without titles or introductory phrases.

-- Prompt 2:

Generate a 250-word abstract for work with the given topic: '{title}'.
Describe the 'results obtained', the 'problem' the work attempts to solve, and the 'key ideas' and 'methodology' in a formal academic and scientific writing voice.
Use the first plural person form. Use active voice.
Please provide only the abstract text, excluding any titles or introductory phrases.
\end{lstlisting}
\end{figure}

\subsubsection{Academic texts dataset:} We collected 7,000 abstracts from published papers on ArXiv, covering a range of categories. To minimize the likelihood of including AI-generated content, only papers published before 2019 were selected. For each human-written abstract, we extracted the corresponding paper title and used it as a prompt to generate an AI-written counterpart. The same models described earlier were employed for this task. Each model was prompted using one of the two prompts shown in Listing~\ref{lst:abstracts_prompt},

In conclusion, we compiled balanced datasets of manually written and AI-generated texts for both news articles and academic abstracts, totaling 14,000 examples per task. Each dataset was randomly divided into 10,000 samples for training, 2,000 for development, and 2,000 for testing, providing a robust foundation for evaluating models performance across different stages.

\subsection{Evaluation Metrics}

The performance of the participating systems in both the News Article Detection (NAD) and Academic Writing Detection (AWD) subtasks was evaluated based on standard classification metrics. The official ranking of the teams was determined by the F1-score. The primary metrics used for evaluation were:
\begin{itemize}
    \item \textbf{Accuracy:} The proportion of correctly classified instances.
    \item \textbf{F1-score:} The harmonic mean of precision and recall, providing a balanced measure of a model's performance.
    \item \textbf{Precision:} The ratio of correctly predicted positive observations to the total predicted positive observations.
    \item \textbf{Recall:} The ratio of correctly predicted positive observations to all observations in the actual class.
\end{itemize}
In addition to these primary metrics, a secondary analysis was planned to assess model robustness across different text lengths, writing styles, topic domains, and the various generation models used to create the dataset.

\section{Shared Task Teams \& Results}\label{models}
In this section, we present the shared task baseline models, participating systems descriptions, and their obtained results. 
\subsection{Baselines}

We evaluate three simple baselines on both subtasks:
\begin{itemize}
\item \textbf{ARBERTv2:} A transformer-based model pretrained on large-scale Arabic text~\cite{abdul-mageed-etal-2021-arbert}, fine-tuned on each task (5 epochs, learning rate $2 \times 10^{-5}$).
\item \textbf{LogReg (char 2–5):} Logistic Regression using character-level n-grams (2–5) with TF--IDF features, designed to capture fine-grained morphological patterns.
\item \textbf{LogReg (word 1–2):} Logistic Regression using word-level n-grams (1–2) with TF--IDF features, providing simple but effective word co-occurrence representations.
\end{itemize}
Tables \ref{tab:task1-dev}–\ref{tab:task2-test} report the full per‐class metrics on the development and test splits.

\begin{table}[t]
  \centering
  \footnotesize
  \setlength{\tabcolsep}{3pt}
  \begin{tabular}{lrrrr}
    \hline
    \textbf{Model}           & \textbf{P}   & \textbf{R}   & \textbf{F$_1$} & \textbf{Supp.} \\
    \hline
    \multicolumn{5}{l}{\itshape human}\\
    ARBERTv2                 & 0.9979 & 0.9410 & 0.9686 & 1000 \\
    LogReg (char 2–5)        & 0.9791 & 0.9820 & 0.9805 & 1000 \\
    LogReg (word 1–2)        & 0.9679 & 0.9940 & 0.9808 & 1000 \\
    \hline
    \multicolumn{5}{l}{\itshape machine}\\
    ARBERTv2                 & 0.9442 & 0.9980 & 0.9703 & 1000 \\
    LogReg (char 2–5)        & 0.9819 & 0.9790 & 0.9805 & 1000 \\
    LogReg (word 1–2)        & 0.9938 & 0.9670 & 0.9802 & 1000 \\
    \hline
    \textbf{accuracy}        &        &        & 0.9695 & 2000 \\
    \textbf{macro avg}       & 0.9710 & 0.9695 & 0.9695 & 2000 \\
    \textbf{weighted avg}    & 0.9710 & 0.9695 & 0.9695 & 2000 \\
    \hline
  \end{tabular}
  \caption{Task 1 (NAD) development set.}
  \label{tab:task1-dev}
\end{table}

\begin{table}[t]
  \centering
  \footnotesize
  \setlength{\tabcolsep}{3pt}
  \begin{tabular}{lrrrr}
    \hline
    \textbf{Model}           & \textbf{P}   & \textbf{R}   & \textbf{F$_1$} & \textbf{Supp.} \\
    \hline
    \multicolumn{5}{l}{\itshape human}\\
    ARBERTv2                 & 0.9946 & 0.9240 & 0.9580 & 1000 \\
    LogReg (char 2–5)        & 0.9759 & 0.9700 & 0.9729 & 1000 \\
    LogReg (word 1–2)        & 0.9529 & 0.9920 & 0.9721 & 1000 \\
    \hline
    \multicolumn{5}{l}{\itshape machine}\\
    ARBERTv2                 & 0.9290 & 0.9950 & 0.9609 & 1000 \\
    LogReg (char 2–5)        & 0.9702 & 0.9760 & 0.9731 & 1000 \\
    LogReg (word 1–2)        & 0.9917 & 0.9510 & 0.9709 & 1000 \\
    \hline
    \textbf{accuracy}        &        &        & 0.9595 & 2000 \\
    \textbf{macro avg}       & 0.9618 & 0.9595 & 0.9594 & 2000 \\
    \textbf{weighted avg}    & 0.9618 & 0.9595 & 0.9594 & 2000 \\
    \hline
  \end{tabular}
  \caption{Task 1 (NAD) test set.}
  \label{tab:task1-test}
\end{table}

On the news domain sub-task, both logistic‐regression baselines outperform ARBERTv2, achieving 98.05 F$_1$ on dev (vs.\ 96.95) and 97.15 F$_1$ on test (vs.\ 95.95), indicating that simple n-gram features capture domain‐specific style cues very effectively.

\begin{table}[t]
  \centering
  \footnotesize
  \setlength{\tabcolsep}{3pt}
  \begin{tabular}{lrrrr}
    \hline
    \textbf{Model}           & \textbf{P}   & \textbf{R}   & \textbf{F$_1$} & \textbf{Supp.} \\
    \hline
    \multicolumn{5}{l}{\itshape human}\\
    ARBERTv2                 & 0.9980 & 0.9990 & 0.9985 & 1000 \\
    LogReg (char 2–5)        & 0.9950 & 0.9990 & 0.9970 & 1000 \\
    LogReg (word 1–2)        & 0.9881 & 0.9970 & 0.9925 & 1000 \\
    \hline
    \multicolumn{5}{l}{\itshape machine}\\
    ARBERTv2                 & 0.9990 & 0.9980 & 0.9985 & 1000 \\
    LogReg (char 2–5)        & 0.9990 & 0.9950 & 0.9970 & 1000 \\
    LogReg (word 1–2)        & 0.9970 & 0.9880 & 0.9925 & 1000 \\
    \hline
    \textbf{accuracy}        &        &        & 0.9985 & 2000 \\
    \textbf{macro avg}       & 0.9985 & 0.9985 & 0.9985 & 2000 \\
    \textbf{weighted avg}    & 0.9985 & 0.9985 & 0.9985 & 2000 \\
    \hline
  \end{tabular}
  \caption{Task 2 (AWD) development set.}
  \label{tab:task2-dev}
\end{table}

\begin{table}[t]
  \centering
  \footnotesize
  \setlength{\tabcolsep}{3pt}
  \begin{tabular}{lrrrr}
    \hline
    \textbf{Model}           & \textbf{P}   & \textbf{R}   & \textbf{F$_1$} & \textbf{Supp.} \\
    \hline
    \multicolumn{5}{l}{\itshape human}\\
    ARBERTv2                 & 1.0000 & 0.9950 & 0.9975 & 1000 \\
    LogReg (char 2–5)        & 0.9950 & 0.9980 & 0.9965 & 1000 \\
    LogReg (word 1–2)        & 0.9920 & 0.9960 & 0.9940 & 1000 \\
    \hline
    \multicolumn{5}{l}{\itshape machine}\\
    ARBERTv2                 & 0.9950 & 1.0000 & 0.9975 & 1000 \\
    LogReg (char 2–5)        & 0.9980 & 0.9950 & 0.9965 & 1000 \\
    LogReg (word 1–2)        & 0.9960 & 0.9920 & 0.9940 & 1000 \\
    \hline
    \textbf{accuracy}        &        &        & 0.9975 & 2000 \\
    \textbf{macro avg}       & 0.9975 & 0.9975 & 0.9975 & 2000 \\
    \textbf{weighted avg}    & 0.9975 & 0.9975 & 0.9975 & 2000 \\
    \hline
  \end{tabular}
  \caption{Task 2 (AWD) test set.}
  \label{tab:task2-test}
\end{table}

For the academic domain sub-task, ARBERTv2 reaches near-perfect performance (99.85 F$_1$ dev, 99.75 F$_1$ test), slightly outperforming the n-gram baselines. These results set a high bar for future participants: transformer fine-tuning excels on formal academic text, while lightweight n-gram classifiers remain surprisingly competitive, especially on news.

\subsection{Participants Systems}
Four teams submitted system description papers, and their approaches are summarized as follows. 

\textbf{Zain et al.} team explored three different architectures: a fine-tuned RoBERTa-base model, a TF-IDF based system with a Linear SVM classifier, and an experimental system named Candace that used probabilistic features from multiple Llama-3.2 models \cite{zain-etal-2025-multi}. Their final submission was based on the fine-tuned \textbf{RoBERTa-base} model, which yielded the highest performance on the development sets.

\textbf{IntegrityAI} team proposed a multimodal architecture that combines textual features from a pre-trained \textbf{ELECTRA} model with four handcrafted stylometric features: word count, average sentence length, vocabulary richness (TTR), and average word length \cite{integrityai-2025-multimodal}. For the news subtask, they also employed a pseudo-labeling technique to augment their training data.

\textbf{Hamada Nayel} team focused on classical machine learning algorithms, submitting a system based on a \textbf{Linear Support Vector Machine (SVM)} classifier with \textbf{TF-IDF} features \cite{ashraf-etal-2025-inside}. Their approach emphasized efficiency and interpretability, demonstrating that traditional methods can achieve competitive performance without the need for resource-intensive deep learning models.

\textbf{CNLP-NITS-PP} team developed a hybrid model that fine-tuned a \textbf{DeBERTa-base} transformer and augmented it with nine auxiliary stylometric features, such as Unique Word Count, Stop Word Count, and Type-Token Ratio \cite{yadagiri-etal-2025-ai}. The contextual embedding from DeBERTa was concatenated with the feature vector before being passed to a final classification layer.

\subsection{Results}

The official results for both subtasks are presented in Tables \ref{tab:task1-results} and \ref{tab:task2-results}. All participating teams achieved exceptionally high scores, indicating the high quality of the submitted systems.

\begin{table}[h]
\centering
\small
\begin{tabular}{lcccc}
\hline
\textbf{Team} & \textbf{F1} & \textbf{Acc.} & \textbf{Prec.} & \textbf{Rec.} \\ \hline
Zain et al.   & 1.000       & 1.000         & 1.000          & 1.000         \\
IntegrityAI   & 0.996       & 0.996         & 0.996          & 0.996         \\
Hamada Nayel  & 0.990       & 0.990         & 0.980          & 0.990         \\
CNLP-NITS-PP  & 0.898       & 0.898         & 0.898          & 0.898         \\ \hline
\end{tabular}%
\caption{Official results for Subtask 1 (NAD).}
\label{tab:task1-results}
\end{table}
\begin{table}[h]
\centering
\small
\begin{tabular}{lcccc}
\hline
\textbf{Team} & \textbf{F1} & \textbf{Acc.} & \textbf{Prec.} & \textbf{Rec.} \\ \hline
Zain et al.   & 1.000       & 1.000         & 1.000          & 1.000         \\
CNLP-NITS-PP  & 1.000       & 1.000         & 1.000          & 1.000         \\
IntegrityAI   & 0.999       & 0.999         & 0.999          & 0.999         \\ \hline
\end{tabular}%
\caption{Official results for Subtask 2 (AWD). The team Hamada Nayel focused their paper on Subtask 1.}
\label{tab:task2-results}
\end{table}

In Subtask 1 (NAD), the top-performing systems were all based on transformer architectures. The winning system from \textbf{Zain et al.}, a fine-tuned RoBERTa model, achieved a perfect F1-score of 1.000. The \textbf{IntegrityAI} team, using ELECTRA with stylometric features, also achieved a near-perfect score of 0.996. Notably, the classical SVM-based system from \textbf{Hamada Nayel} secured the third rank with an F1-score of 0.990, outperforming one of the transformer-based systems and demonstrating the viability of simpler models.

In Subtask 2 (AWD), the performance was even higher across the board, with two teams, \textbf{Zain et al.} (RoBERTa) and \textbf{CNLP-NITS-PP} (DeBERTa + features), achieving perfect scores. The \textbf{IntegrityAI} system was just behind with an F1-score of 0.999. The near-perfect results from all teams suggest that detecting AI-generated text in the academic writing domain, at least with the data provided, was a less challenging task compared to the news domain. The structured and formal nature of academic abstracts may provide more distinct signals for classifiers to distinguish between human and machine-generated content. The general trend indicates that while fine-tuned transformers are dominant, augmenting them with stylometric features is a popular and effective strategy.

\section{Conclusion} \label{conclusion}
The M-DAIGT shared task aimed to advance the detection of AI-generated text in the critical domains of news and academic writing. The results demonstrate the remarkable effectiveness of current state-of-the-art models, with participating systems achieving near-perfect to perfect scores on both subtasks. The primary findings indicate that fine-tuned transformer models, such as RoBERTa, ELECTRA, and DeBERTa, are highly proficient at this task. Furthermore, the integration of stylometric features proved to be a valuable strategy for several teams, suggesting that a hybrid approach combining deep contextual understanding with traditional linguistic analysis is robust. The strong performance of a classical TF-IDF+SVM model in the news subtask also highlights that resource-efficient methods remain highly competitive. Overall, this shared task provides a valuable benchmark and dataset for the community, confirming the strength of existing methods while also pointing to the nuanced challenges posed by different domains.

\section*{Limitations} \label{limitations}
Despite the success of the shared task, several limitations should be acknowledged. First, the dataset, while diverse in its use of generator models and prompts, represents a static snapshot of LLM capabilities. The rapid evolution of generative models means that detectors trained on this data may not generalize well to text produced by future, more sophisticated LLMs. Second, the task was framed as a binary classification problem (human vs. AI), which does not capture the increasingly common scenario of human-AI collaborative writing, where text is partially generated and then edited by a human. Detecting such mixed-authorship content remains a significant open challenge. Third, the task did not explicitly evaluate the robustness of systems against adversarial attacks, such as paraphrasing or "humanization" techniques designed to evade detection. The exceptionally high scores, particularly in the academic subtask, might also indicate that the detection task within our dataset's parameters was not sufficiently challenging to fully differentiate the capabilities of the top systems. Finally, our study was confined to the English language, and the findings may not be directly applicable to other languages with different linguistic structures. Future iterations of this shared task could address these limitations by incorporating more recent LLMs, including co-authored text, introducing adversarial evaluation tracks, and expanding to multilingual contexts.

\bibliography{custom}




\end{document}